\documentclass[lettersize,journal]{IEEEtran}
\usepackage{amsmath,amsfonts}
\usepackage{algorithmic}
\usepackage{algorithm}
\usepackage{array}
\usepackage[caption=false,font=normalsize,labelfont=sf,textfont=sf]{subfig}
\usepackage{textcomp}
\usepackage{stfloats}
\usepackage{url}
\usepackage{verbatim}
\usepackage{graphicx}
\usepackage{cite}
\usepackage{xcolor}
\usepackage{tabularx}

\usepackage{amsmath}
\usepackage{amssymb}
\usepackage{multirow}
\usepackage{booktabs}
\usepackage{colortbl}
\graphicspath{{./Image/}}

\begin{document}

\title{Diff-RF: Mutually Reinforced Image Registration and Fusion via Degradation-Aware Learning}

\author{Xunpeng Yi, Zaixi Du, Qinglong Yan, Yibing Zhang, Han Xu, and Jiayi Ma
\thanks{This work was supported by the National Natural Science Foundation of China under Grant nos. 625B2135 and 62676289. \emph{(Corresponding author: Jiayi Ma.)}}
\thanks{Xunpeng Yi, Zaixi Du, Qinglong Yan and Yibing Zhang are with the Electronic Information School, Wuhan University, Wuhan 430072, China (e-mail: yixunpeng@whu.edu.cn, zaixidu@whu.edu.cn, qinglong\_yan@whu.edu.cn, zhangyibing@whu.edu.cn).}
\thanks{Han Xu is with the School of Automation, Southeast University, Nanjing 210096, China (e-mail: xu\_han@seu.edu.cn).}
\thanks{Jiayi Ma is with the Electronic Information School and the School of Robotics, Wuhan University, Wuhan, 430072 (e-mail: jyma2010@gmail.com)}
}

\markboth{}%
{Yi \MakeLowercase{\textit{et al.}}: Title}


\maketitle

\begin{abstract}
Image registration and fusion aim to establish spatial correspondences from misaligned multi-modal source images, and integrate complementary information. However, in real-world imaging scenarios, source images are often affected by complex and diverse degradations, such as low illumination, noise, etc., which severely hinder the effectiveness of registration and fusion. To address this issue, we propose a mutually reinforced image registration and fusion diffusion framework via degradation-aware learning, termed Diff-RF. It explores the intrinsic coupling between registration–fusion and information restoration in the degradation conditions, enabling high-quality fusion of unregistered images under complex degradation conditions. First, the intra-modal restoration module is designed to alleviate modality-specific degradations by leveraging information within each modality, thereby providing more reliable structural representations for registration and facilitating subsequent cross-modal fusion. Second, we develop a cross-modal diffusion registration and fusion module that establishes bidirectional interaction between registration and fusion. By integrating fusion-derived visual cues and correspondence-based geometric conditions into the diffusion process, the proposed framework progressively refines spatial alignment and exploits cross-modal complementary information to achieve collaborative enhancement. Rather than treating them as independent components, degradation-aware information restoration and the collaborative optimization of registration and fusion are tightly coupled, achieving overall performance improvements. Extensive experiments on multiple extended datasets demonstrate that Diff-RF achieves superior registration accuracy and fusion quality under various degraded scenarios, exhibiting strong robustness and generalization ability. The code is available at https://github.com/XunpengYi/Diff-RF.
\end{abstract}


\section{Introduction}
\IEEEPARstart{I}{mage} fusion is a fundamental task in image processing, aiming to integrate the images from different sources into a unified representation fusion image that comprehensively preserves scene information~\cite{ma2020ddcgan, cao2023multi, he2023degradation, yi2024diffif}. It can overcome the limitations imposed by the restricted information sources of single-modality images, and has therefore been widely applied in autonomous driving, industrial inspection, and community security~\cite{sun2022detfusion, yi2025artificial, wu2025every}.

Infrared and visible image fusion aims to generate a unified and comprehensive scene representation by integrating the rich texture details of visible images with the thermal radiation information captured by infrared images. Nevertheless, differences in the positions and viewpoints of multi-modal imaging devices inevitably introduce parallax between the acquired source images~\cite{xu2022rfnet, wang2022unsupervised, li2025mulfs}. The resulting pixel-level misalignment can impair the fusion process, causing artifacts and structural distortions in the fused images, which may further degrade the performance of downstream tasks.

\begin{figure}[!t]
\centering
\includegraphics[width=0.95\linewidth]{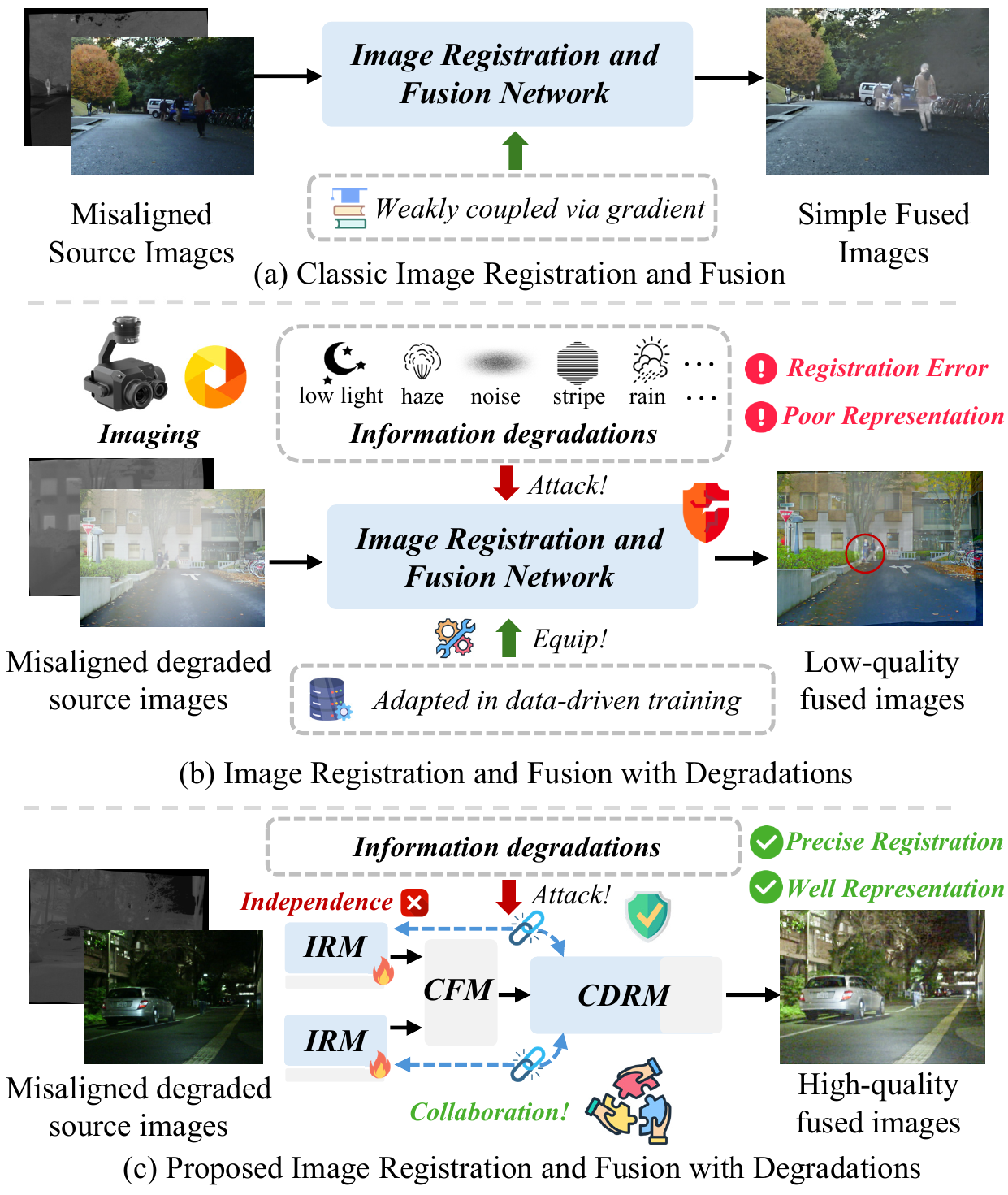}
\caption{Toward the image registration and fusion with degradations in complex scenarios. (a) Classic image registration and fusion frameworks. (b) Data-driven registration–fusion frameworks with low adaptation to degradations. (c) Proposed mutually reinforced image
registration and fusion via degradation-aware learning. }
\label{fig1-1}
\end{figure}

Recent image fusion methods have increasingly recognized that degradations arising during the imaging process can adversely affect the visual quality and information representation of fused images~\cite{yi2024textif, tang2024drmf}. These methods typically introduce supervision through various strategies to suppress degradations during the fusion process. However, this paradigm generally relies on the assumption that the source images are well aligned. Therefore, they overlook the fact that imaging degradations may already be present in the prerequisite registration stage before fusion, thereby affecting both registration accuracy and the implementation of registration–fusion integration.

\begin{figure*}[!t]
\centering
\includegraphics[width=1\linewidth]{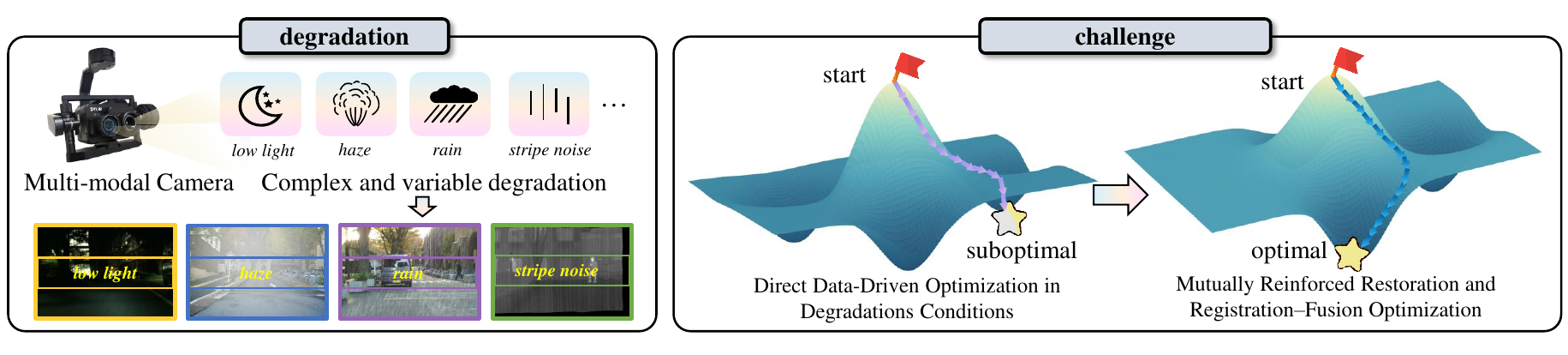}
\caption{Illustration of degradation in real-world infrared-visible multi-modal imaging and the local optimization issue caused by direct data-driven learning.}
\label{fig2-1}
\end{figure*}

This further raises a critical question: how can a model effectively perform registration and fusion under degraded imaging conditions? Merely relying on a data-driven strategy to force the feature extractor to adapt to degradations cannot fundamentally resolve this issue, as shown in Fig.~\ref{fig1-1}. They still rely on feature extractors to implicitly learn degradation-robust registration features from degraded observations, which significantly increases optimization difficulty, as illustrated in Fig.~\ref{fig2-1}, and may entangle degradation factors with useful structural information, making it difficult to satisfy the requirements of accurate registration and high-quality fusion. A key challenge, therefore, lies in decoupling task-relevant representations from degraded source images and establishing unified features that can support registration and fusion.

To address this problem, we propose a degradation-aware image registration and fusion diffusion framework, termed Diff-RF. It explores the intrinsic coupling among restoration, cross-modal correspondence estimation, and information fusion, establishing a unified framework for mutually reinforced registration and fusion. In detail, considering the intrinsic relationship between degradations and registration–fusion, we integrate intra-modal enhancement with bidirectional mutual feedback between cross-modal registration and fusion. Rather than treating restoration, registration, and fusion as independent sequential components, Diff-RF tightly couples the intra-modal restoration module, the cross-modal diffusion registration and fusion module, enabling their joint optimization and mutual reinforcement. The intra-modal restoration module alleviates modality-specific degradations by exploiting inherent information within each modality, thereby generating more reliable structural representations for correspondence estimation and facilitating subsequent fusion. The cross-modal diffusion registration and fusion module integrates fusion-derived visual cues and correspondence-based geometric conditions obtained from restoration into the diffusion registration process, enabling fusion-guided registration. Furthermore, fusion is performed based on the deformation field estimated by registration, thereby establishing a bidirectional interaction between registration and fusion. Through the mutual reinforcement between degradation compensation, spatial alignment, and cross-modal enhancement, Diff-RF achieves robust registration and high-quality fusion under complex degraded scenarios.

Overall, our contributions can be summarized as follows:
\begin{itemize}
  \item[-] To adapt registration and fusion to complex scenarios, we propose a novel degradation-aware framework for jointly addressing image registration and fusion. Specifically, we decouple the restoration task from registration and fusion, and further enable their mutual enhancement.
  \item[-] We reveal the insight that restoration should precede registration and work with it for degraded multi-modal images, as modality-specific degradations hinder reliable correspondence estimation. Accordingly, we develop an intra-modal restoration module jointly optimized with registration–fusion to establish reliable structural and geometric representations, thereby facilitating robust registration and fusion.
  \item[-] We establish a coupling between fusion and registration by explicitly feeding fused representations into the registration, providing discriminative visual cues for cross-modal correspondence estimation. Building on this, the diffusion progressively models spatial relationships and iteratively refines registration conditioned on the current cross-modal misalignment, forming a collaborative mechanism in which fusion facilitates spatial estimation and misalignment guides registration refinement.
  \item[-] Extensive experiments demonstrate that the proposed method achieves superior visual quality and registration performance across multiple extended datasets, while effectively suppressing various complex degradations and exhibiting strong robustness and generalization capability.

\end{itemize}

\section{Related Work}
\subsection{Classical Image Fusion Methods}
Early image fusion methods mainly relied on hand-crafted fusion strategies to integrate complementary information from multi-modal source images. With the rapid development of deep learning, the field of image fusion has witnessed substantial progress. Among these advances, autoencoder-based methods first achieved effective information aggregation by leveraging pre-trained encoder–decoder architectures and carefully designed fusion rules~\cite{liu2016image, li2018densefuse}. Subsequently, end-to-end optimization frameworks driven by fusion-specific loss functions were introduced~\cite{xu2020u2fusion}, further improving the flexibility and representation capability of image fusion models. Furthermore, image fusion methods based on various advanced architectures have been extensively explored, including GAN-based frameworks~\cite{ma2020ddcgan, zhang2023transformer}, Transformer-based models~\cite{zhao2023cddfuse, yi2024textif}, diffusion-based architectures~\cite{zhao2023ddfm, yi2024diffif, yang2025lfdt}, and lookup-table-based approaches~\cite{yi2025lut}. These developments have substantially promoted the progress of image fusion, particularly in terms of fusion performance and computational efficiency. Meanwhile, another line of methods integrates image fusion with downstream tasks, such as object detection~\cite{liu2022target, sun2022detfusion}. By adopting cascaded or task-driven paradigms, these methods promote the improvement of the semantic performance of image fusion.

\subsection{Image Registration and Fusion Methods}
Due to differences in the positions and viewing angles of multi-modal imaging devices, the acquired source images often suffer from parallax and pixel-level misalignment. To address this problem, RFNet performs image registration by introducing image style transfer and a coarse-to-fine registration strategy~\cite{xu2022rfnet}. Moreover, it leverages the sparsity of fusion gradients to enable mutual promotion between image registration and fusion. Building upon this, MURF optimizes feature extraction through contrastive learning and achieves better performance~\cite{xu2023murf}. Correspondingly, UMF-CMGR simultaneously accomplishes registration and fusion by combining a GAN-based transformation method with multi-level fusion~\cite{wang2022unsupervised}. AUNet further upgrades the coupling between fusion and registration to the feature level, and leverages the powerful modeling capability of diffusion models for transformation, achieving improved performance~\cite{lu2025net}. FusionRegister adopts a fusion-first paradigm, where image fusion is first conducted and the fused results are subsequently refined to achieve more accurate registration in salient regions~\cite{bian2026fusionregister}. Despite the remarkable progress achieved by these methods, recent studies have shown that degradations in fused images can severely affect registration quality. Nevertheless, existing methods rarely provide an essential discussion on the intrinsic relationship between fusion and registration under complex degraded scenarios.

\begin{figure*}[!t]
\centering
\includegraphics[width=1\linewidth]{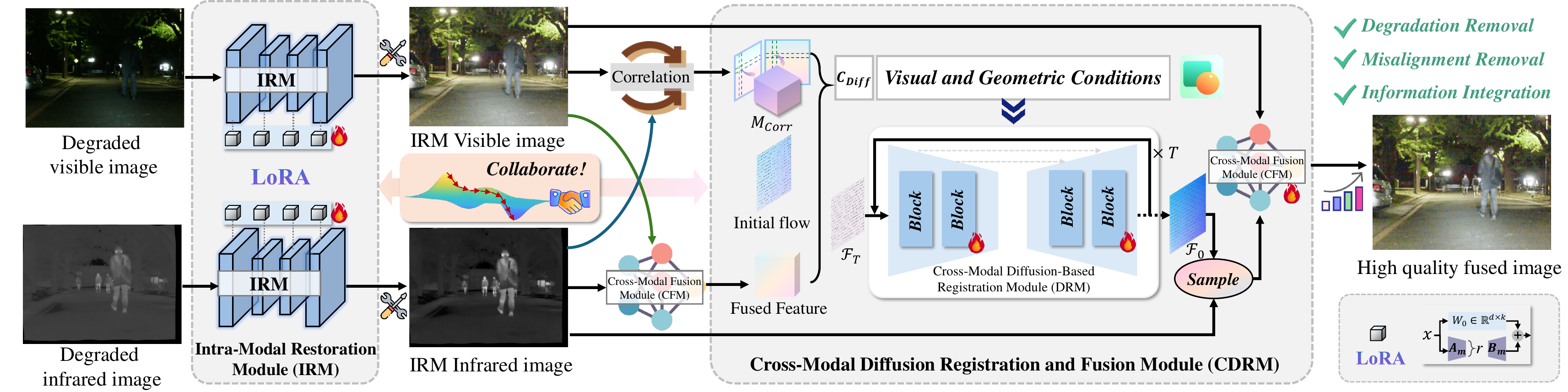}
\caption{The framework of Diff-RF. It explores the intrinsic coupling among intra-modal restoration, cross-modal correspondence estimation, and information fusion, forming a unified framework with mutual reinforcement between registration and fusion.}
\label{fig3-1}
\end{figure*}

\section{Methodology}
In this section, we first present the problem formulation and introduce underlying principles. As illustrated in Fig.~\ref{fig3-1}, we subsequently describe the overall framework of Diff-RF, which consists of an Intra-Modal Restoration Module, a Cross-Modal Diffusion Registration and Fusion Module. Finally, we detail the loss functions for model optimization.

\subsection{Problem Formulation}
Given unregistered and degraded source images captured under complex imaging conditions $\{I_{vis}^{D}, I_{ir}^{D}\}$, we develop an iterative and mutually reinforcing framework for registration and fusion. In detail, the overall process is decomposed into the intra-modal restoration and cross-modal diffusion-based registration and fusion.

Intra-modal restoration leverages the observation information within each modality to recover modality-specific degradations, thereby providing higher-quality and structurally more stable image representations for subsequent cross-modal information interaction.
\begin{equation}
I_{vis}^{IRM}=\theta_{IRM-vis}^{LoRA}(I_{vis}^{D}),\quad I_{ir}^{IRM}=\theta_{IRM-ir}^{LoRA}(I_{ir}^{D}),
\label{eq1}
\end{equation}
where $\theta_{IRM-vis}^{LoRA}$ and $\theta_{IRM-ir}^{LoRA}$ denote the parameters of the pretrained networks that are further fine-tuned with LoRA through joint optimization with the registration–fusion.
After the intra-modal degradations are preliminarily mitigated, the spatial and structural correspondences across different modalities become more explicit, facilitating more accurate cross-modal registration. Building upon this, cross-modal diffusion-based registration and fusion further exploits the complementary information between modalities. By leveraging the coupling between registration and fusion, it jointly promotes spatial alignment and information integration, while further improving the restoration quality:
\begin{equation}
Flow=\theta_{Reg-Diff}^{T}(Flow_{init},M_{Corr},F_{fus}),
\label{eq2}
\end{equation}

\begin{equation}
F_{fus}=\mathcal{E}_{reg}(\theta_{fus}(I_{vis}^{IRM},I_{ir}^{IRM})),
\label{eq3}
\end{equation}
where $Flow$ and $Flow_{init}$ denote the estimated flow field and the initial flow field used for the diffusion network $\theta_{Reg-Diff}^{T}$, respectively. $M_{Corr}$ denotes the estimated cross-modal correspondence matrix. $\mathcal{E}_{reg}$ is the feature extractor used for image registration. $\theta_{fus}$ represents the fusion network.

Subsequently, the flow field estimated by the diffusion is used to warp the intra-modally restored images, followed by further cross-modal fusion and collaborative restoration by exploiting complementary information across modalities:
\begin{equation}
\widetilde{I}_{ir}^{IRM}=warp(I_{ir}^{IRM},Flow),
\label{eq4}
\end{equation}
\begin{equation}
I_{f}=\theta_{fus}(\widetilde{I}_{ir}^{IRM}, I_{vis}^{IRM}),
\label{eq5}
\end{equation}
where $I_{f}$ represents the final fused output, which integrates cross-modal information fusion and restoration. This collaborative framework of intra-modal restoration, cross-modal registration, and multi-modal fusion effectively adapts the registration–fusion operation to the complex degraded scenarios.

\subsection{Intra-Modal Restoration Module (IRM)}
Considering the adverse effects of inherent imaging degradations on subsequent processing, we first perform intra-modal restoration to obtain reliable structural and representational information for subsequent registration and fusion. Specifically, the infrared and visible modalities are restored in a decoupled manner using relatively lightweight backbones. Simply, it can be formulated as:
\begin{equation}
I_{vis}^{IRM}=\theta_{IRM-vis}(I_{vis}^{D}),\quad I_{ir}^{IRM}=\theta_{IRM-ir}(I_{ir}^{D}),
\label{eq6}
\end{equation}
where $\theta_{IRM-vis}$ and $\theta_{IRM-ir}$ are parameters of restoration network. Although image enhancement and registration share a positive transfer relationship through their common demand for high-quality inputs, close interaction between the two tasks is still required to achieve mutual improvement. To this end, we replace the independently cascaded paradigm with a registration-coupled fine-tuning strategy and introduce Low-Rank Adaptation (\textit{LoRA})~\cite{hu2022lora} into the IRM. This design enables the restoration module to recover high-quality information representations while simultaneously learning registration-friendly structural recovery. It can be formulated as:
\begin{equation}
y=\mathbf{W}_{0}x+\frac{\alpha}{r}\mathbf{B}_{m}\mathbf{A}_{m}x,\quad m\in\{ir, vis\},
\label{eq7}
\end{equation}
where $x$ and $y$ denote the input and output of the projection. Equivalently, the effective projection weight can be expressed as:
\begin{equation}
\mathbf{W}_{m}^{eff}=\mathbf{W}_{0}+\Delta\mathbf{W}_{m},
\label{eq8}
\end{equation}
\begin{equation}
\Delta\mathbf{W}_{m}=\frac{\alpha}{r}\mathbf{B}_{m}\mathbf{A}_{m},
\label{eq9}
\end{equation}
where $\mathbf{A}_{m}$ and $\mathbf{B}_{m}$ are modality-specific low-rank projections. $r$ denotes the low-rank dimension. $\alpha$ controls the adaptation strength. During adaptation, we optimize the projection convolutions within each block, while keeping the original pretrained enhancement weights $\mathbf{W}_{0}$ frozen. Specifically, only the LoRA matrices $\mathbf{A}_{m}$ and $\mathbf{B}_{m}$ are updated, with $\mathbf{B}_{m}$ initialized to zero such that the initial weight increment $\Delta\mathbf{W}_{m}$ is zero, thereby ensuring stable optimization at the beginning of training.

Accordingly, the IRM is parameterized by the frozen pretrained restoration parameters and the learnable registration-oriented adaptation. For each modality $m\in\{ir,vis\}$, the general IRM formulation in Eq.~\ref{eq1} can be instantiated as:
\begin{equation}
\theta_{IRM-m}^{LoRA}(I_{m}^{D})=\mathcal{R}(I_{m}^{D};\theta_{IRM-m},\Delta\theta_{LoRA-m}),
\label{eq10}
\end{equation}
where $\theta_{IRM-m}$ and $\Delta\theta_{LoRA-m}$ are the frozen pretrained restoration parameters and learnable registration-oriented adaptation, respectively.

\begin{figure}[!t]
\centering
\includegraphics[width=1\linewidth]{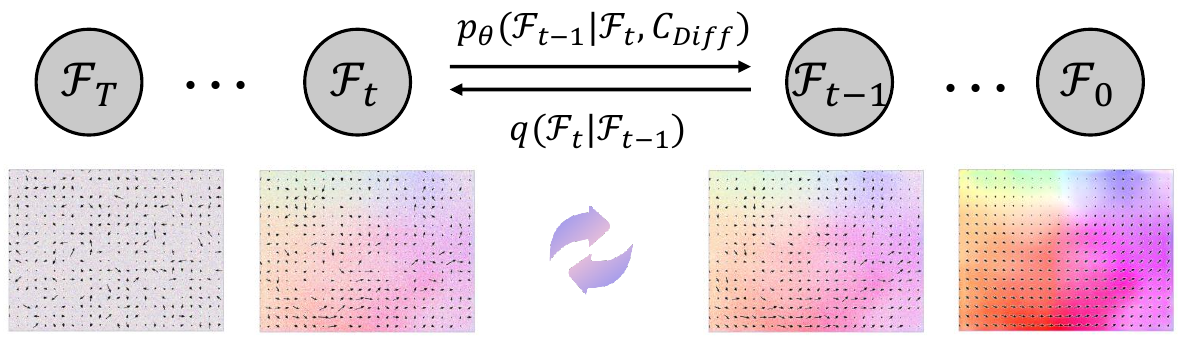}
\caption{Visual progressive diffusion process of the flow in CDRM.}
\label{fig4-1}
\end{figure}

\subsection{Cross-Modal Diffusion Registration and Fusion Module (CDRM)}
After obtaining reliable structural and representational information through IRM, establishing cross-modal registration and fusion relationships while further mitigating degradations becomes crucial. Cross-Modal Diffusion Registration and Fusion Module aims to establish a coupled registration–fusion framework, where registration and fusion mutually reinforce each other to achieve better overall performance, while cross-modal information is further exploited to alleviate degradations. We design CDRM
composed with a cross-modal diffusion-based registration module (DRM) and a cross-modal fusion module (CFM).

Different from existing fusion gradient-guided strategies, the fused features are directly incorporated into the registration to provide information about the degree of misalignment. We first perform preliminary fusion on the unregistered images:
\begin{equation}
I_{fus}^{mis}=\theta_{fus}(I_{vis}^{IRM},I_{ir}^{IRM}).
\label{eq11}
\end{equation}
Due to the spatial misalignment between the source images, noticeable artifacts inevitably appear in the fused image. The registration feature extractor is employed to extract fused features from the preliminarily fused but unregistered image, which are then used to guide the subsequent registration:
\begin{equation}
F_{fus}=\mathcal{E}_{reg}(I_{fus}^{mis}),
\label{eq12}
\end{equation}
where $\mathcal{E}_{reg}$ denotes the registration feature extractor. $F_{fus}$ provides fusion-derived cues that implicitly characterize the spatial inconsistency between the two modalities. Also, reliable diffusion-based registration further requires an explicit geometric prior to characterize cross-modal correspondences. Therefore, in parallel with the fusion-derived condition, we construct a correspondence-derived geometric condition from the restored modality-specific representations. Specifically, the cross-modal correlation is first computed as:
\begin{equation}
M_{Corr}=Corr(\mathcal{E}_{reg}(I_{vis}^{IRM}), \mathcal{E}_{reg}(I_{ir}^{IRM})),
\label{eq13}
\end{equation}
where $I_{ir}^{IRM}$ and $I_{vis}^{IRM}$ denote the infrared and visible images restored by IRM, respectively. And $Corr$ represents the cross-modal correlation operation. Subsequently, the correlation representation is transformed into an initialized flow field to provide more direct flow-related information as a condition:
\begin{equation}
Flow_{init}=\mathcal{T}_{M\xrightarrow{}F}(M_{Corr}),
\label{eq14}
\end{equation}
where $\mathcal{T}_{M\xrightarrow{}F}$ denotes the correlation-to-flow transformation. In this way, $F_{fus}$, $M_{Corr}$, and $Flow_{init}$ serve as fundamental registration conditions, where the visual features in $F_{fus}$ capture fusion-derived misalignment cues, while the geometric information from $M_{Corr}$ and $Flow_{init}$ provides an explicit initialization of cross-modal correspondence.

Built on this, we formulate cross-modal registration as a conditional diffusion model, as shown in Fig.~\ref{fig4-1}. According to the Markov chain property and the definition of diffusion models, we can derive the forward diffusion:
\begin{equation}
q(\mathcal{F}_{t}|\mathcal{F}_{0})=\mathcal{N}(\mathcal{F}_{t}; \sqrt{\overline{\alpha}_{t}}\mathcal{F}_{0}, (1-\overline{\alpha}_{t})\mathbf{I}),
\label{eq15}
\end{equation}
where $\mathcal{F}_{t}$ and $\mathcal{F}_{0}$ denote the flow field at $t$ step and ground-truth flow field. $\alpha_{t}$ represents the cumulative noise schedule coefficient. As the diffusion proceeds, the flow field $\mathcal{F}_{T}$ gradually approaches a pure Gaussian distribution. The reverse sampling process aims to recover the complete multi-modal registration flow field by leveraging the provided visual fusion and geometric alignment conditions. Accordingly, the optimization procedure can be formulated as:
\begin{equation}
\min \|\epsilon-\epsilon_{\theta}(\sqrt{\overline{\alpha}_{t}} \mathcal{F}_{0}+\sqrt{1-\overline{\alpha}_{t}}\epsilon, C_{Diff}, t)\|,
\label{eq16}
\end{equation}
where $C_{Diff}=\{F_{fus}, M_{corr}, Flow_{init}\}$ is the diffusion visual fusion and geometric alignment conditions. $\epsilon$ and $\epsilon_{\theta}$ denote the diffusion noise and noise prediction network. By establishing the relationship between the noise and the estimated flow field, the above optimization objective can be reformulated as an approximate direct constraint on the flow field predicted by the network:
\begin{equation}
\widetilde{\mathcal{F}}_{0,t}=\frac{1}{\sqrt{\overline{\alpha}_{t}}}(\mathcal{F}_{t}-\sqrt{1-\overline{\alpha}_{t}}\epsilon_{\theta}(\mathcal{F}_{t}, C_{Diff}, t)),
\label{eq17}
\end{equation}
\begin{equation}
\min \|\mathcal{F}_{0}-\widetilde{\mathcal{F}}_{0,t}\|=\|\mathcal{F}_{0}-\theta_{Reg-Diff}^{T}(\mathcal{F}_{t}, C_{Diff}, t))\|,
\label{eq18}
\end{equation}
where $\mathcal{F}_{0}$ is the ground-truth flow field. Through the iterative process of cross-modal diffusion registration, the final output flow field $Flow=\widetilde{\mathcal{F}}_{0}$ is obtained under the guidance of visual fusion and geometric correspondence information. Rather than treating registration as an isolated step, the estimated flow field establishes the geometric consistency required for subsequent cross-modal aggregation. Specifically, it transforms the restored modality-specific representations into a spatially aligned feature space, where complementary information can be effectively integrated:
\begin{equation}
\widetilde{I}_{ir}^{IRM}=warp(I_{ir}^{IRM},Flow),
\label{eq19}
\end{equation}
\begin{equation}
I_{f}=\theta_{fus}(\widetilde{I}_{ir}^{IRM}, I_{vis}^{IRM}).
\label{eq20}
\end{equation}
Beyond the independent intra-modal restoration, the subsequent registration-guided fusion further enables cross-modal enhancement by jointly leveraging the complementary information from the aligned infrared and visible representations. Consequently, CDRM does not perform registration and fusion sequentially, but forms a bidirectional registration–fusion coupling paradigm. fusion-derived information facilitates diffusion-based registration, while registration-induced geometric consistency enables cross-modal collaborative enhancement. This interaction allows the model to simultaneously improve spatial alignment and restoration quality under complex degradation conditions.

\subsection{Loss Functions}
For the Intra-Modal Restoration Module, we optimize the restoration objective by enforcing an L1 reconstruction loss between the restored images and their corresponding ground-truth observations:
\begin{equation}
L_{IRM}=\sum_{m\in\{vis,ir\}}\left\|I_{m}^{GT}-I_{m}^{IRM}\right\|,
\label{eq21}
\end{equation}
where $I_{m}^{GT}$ and $I_{m}^{IRM}$ denote the ground-truth image and the restoration result predicted by IRM for modality $m$, respectively.

For the Cross-Modal Diffusion Registration and Fusion Module, we supervise the cross-modal geometric correspondence in DRM by minimizing the mean squared error between the ground-truth flow field and the flow estimated through the diffusion denoising process:
\begin{equation}
L_{Reg}=\|\mathcal{F}_{0}-\widetilde{\mathcal{F}}_{0,t}\|_{2}^{2} + \lambda_{r}L_{smooth},
\label{eq22}
\end{equation}
where $\mathcal{F}_{0}$ and $\widetilde{\mathcal{F}}_{0,t}$ denote the ground-truth flow field and the estimated flow field generated by the diffusion model at the $t$-th step, respectively. $L_{smooth}$ denotes the smoothness regularization term for stable flow estimation. Also,  we optimize the cross-modal fusion and enhancement of CFM using a fusion loss, which consists of intensity, gradient, color, and structural similarity losses:
\begin{equation}
L_{Fus}=L_{int}+\alpha_{grad}L_{grad}+\alpha_{color}L_{color}+\alpha_{ssim}L_{ssim}.
\label{eq23}
\end{equation}

Among them, $L_{int}$ encourages the fused image to preserve salient thermal target information from the infrared modality while maintaining the overall scene intensity:
\begin{equation}
L_{int}=\frac{1}{HW}\|I_{f}-max(I_{vis}^{GT},I_{ir}^{GT})\|_{1}.
\label{eq24}
\end{equation}
The gradient loss $L_{grad}$ is introduced to preserve complementary high-frequency details by enhancing the gradient consistency between the fused output and the ground-truth source images:
\begin{equation}
L_{grad}=\frac{1}{HW}\|\nabla I_{f}-max(\nabla I_{vis}^{GT},\nabla I_{ir}^{GT})\|_{1},
\label{eq25}
\end{equation}
where $\nabla$ is the gradient operator. Furthermore, to maintain natural color appearance, a color consistency loss $L_{color}$ is designed in the CbCr color space:
\begin{equation}
L_{color}=\frac{1}{HW}\|\mathcal{F}_{CbCr}(I_{f})-\mathcal{F}_{CbCr}(I_{vis}^{GT})\|_{1},
\label{eq26}
\end{equation}
where $\mathcal{F}_{CbCr}$ denotes the transfer function of RGB to CbCr space. Moreover, the structural similarity loss $L_{ssim}$ is employed to enhance structural fidelity and perceptual consistency between the fused image and the source modalities:
\begin{equation}
L_{ssim}=1-SSIM(I_{f},I_{vis}^{GT})+1-SSIM(I_{f},I_{ir}^{GT}).
\label{eq27}
\end{equation}

On the basis of this, these loss objectives are jointly optimized to train the proposed registration-fusion framework. The overall training objective is formulated as:
\begin{equation}
L_{total}=L_{IRM}+\alpha_{Reg}L_{Reg}+\alpha_{Fus}L_{Fus},
\label{eq28}
\end{equation}
where $\alpha_{Reg}$ and $\alpha_{Fus}$ denote the weighting hyperparameters for balancing different loss terms.

\begin{figure*}[!t]
\centering
\includegraphics[width=1\linewidth]{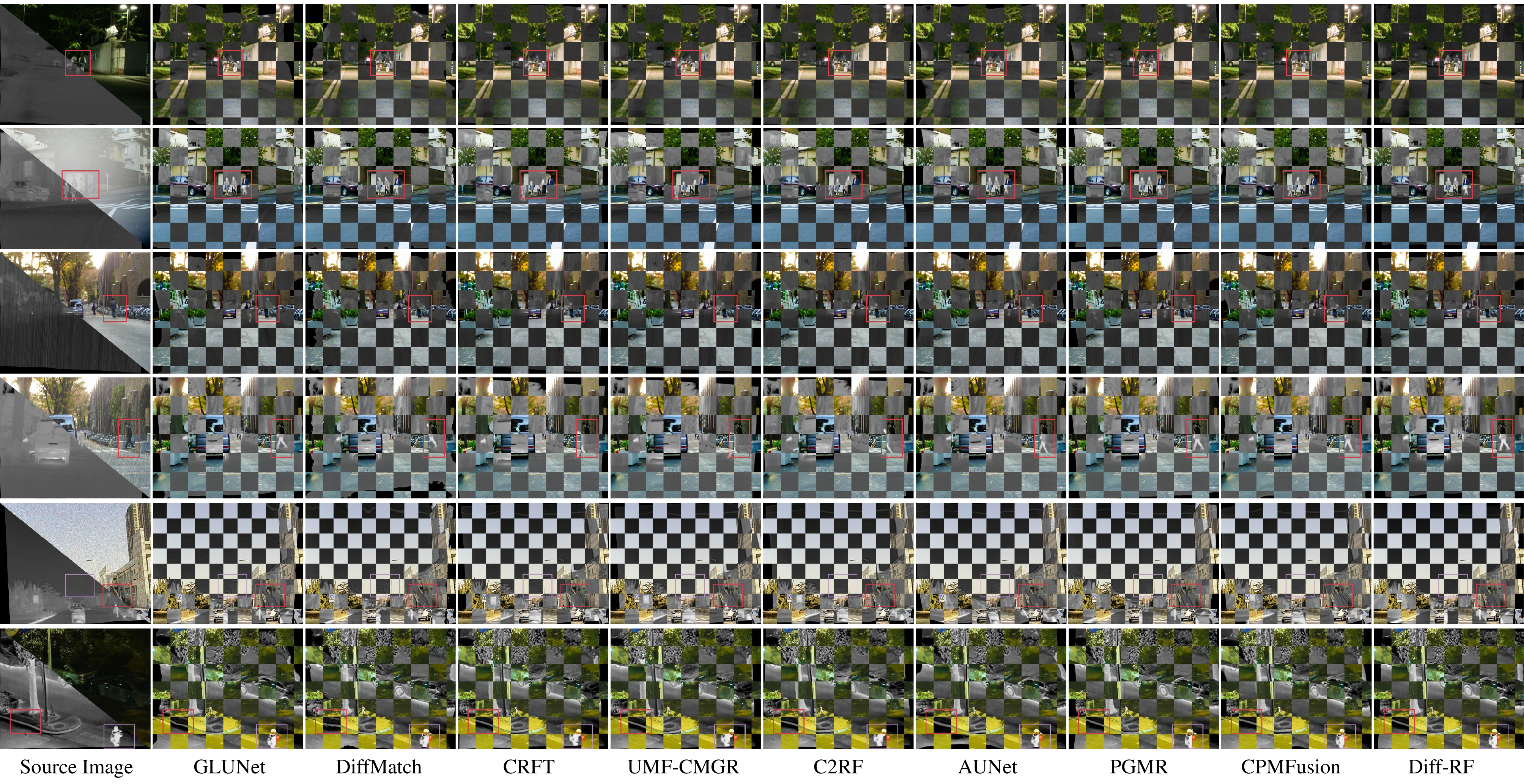}
\caption{Qualitative comparison on the MFNet dataset with state-of-the-art image registration and fusion methods. ``Reg.'' indicates that registration preprocessing is performed using the best-performing registration method on the validation set.}
\label{reg_qual}
\end{figure*}

\section{Experiments}
In this section, we first introduce the implementation details and experimental settings. Subsequently, we conduct a series of qualitative and quantitative evaluations. Finally, we perform downstream-task experiments and ablation studies to demonstrate the practical applicability of the proposed method and validate the contribution of each component.

\noindent
\textbf{Implementation Details:} We adopt a two-stage optimization strategy to train Diff-RF. First, the Intra-Modal Restoration Module and the Cross-Modal Diffusion Registration and Fusion Module are optimized without gradient sharing for 180K iterations to ensure stable learning of their respective objectives. AdamW~\cite{loshchilov2019decoupled} is adopted as the optimizer with a learning rate of $1\times10^{-5}$ and a batch size of 4. The images are randomly cropped to a resolution of $520\times520$ during training. Subsequently, the main parameters of IRM are frozen to maintain its single-modal restoration capability, while LoRA is jointly optimized with CDRM to enable collaborative optimization between restoration and registration–fusion. Similarly, we employ the AdamW optimizer with a learning rate of $5\times10^{-6}$ and optimize the model for 40K iterations. The batch size is set to 2. All experiments are implemented using the PyTorch~\cite{paszke2019pytorch} framework on a machine equipped with the NVIDIA RTX 3090 GPUs.

\noindent
\textbf{Datasets and Metric:} To comprehensively validate the effectiveness of the proposed method, we conduct extensive experiments on several widely used infrared and visible image datasets, including MFNet~\cite{ha2017mfnet}, FMB~\cite{liu2023multi}, and LLVIP~\cite{jia2021llvip} datasets. All these datasets are collected under real-world conditions and naturally contain certain inherent degradations. In particular, we introduce additional simulated degradations into MFNet and FMB to expand the diversity of degradation scenarios. For the MFNet dataset, we use 1996 image pairs for training and 786 image pairs for testing. For the FMB dataset, 3074 image pairs are utilized for training, while 703 image pairs are reserved for evaluation. Correspondingly, LLVIP is directly used without additional processing to evaluate the generalization capability under real-world conditions. Specifically, we use 180 infrared–visible image pairs from LLVIP for generalization evaluation. To simulate real-world misregistration, following previous works~\cite{xu2023murf, tang2025c2rf, lu2025net, xiao2026cpmfusion}, we apply random geometric transformations to \([-3.84^\circ, 3.84^\circ]\), translations \([-32, 32]\) and \([-25.6, 25.6]\) pixels along the horizontal and vertical directions, respectively, and smooth elastic deformations with displacement fields of up to 5 pixels per component.

\noindent
For the registration task, we employ the endpoint error (EPE)~\cite{baker2011database} and the estimation accuracy under 1-pixel, 3-pixel, and 5-pixel thresholds as evaluation metrics to quantify the discrepancy between the predicted and ground-truth flow fields. These metrics comprehensively evaluate registration performance by considering both the magnitude of flow estimation errors and the correspondence accuracy under different precision thresholds. A lower EPE value indicates a smaller error between the predicted and ground-truth flow fields, reflecting higher registration accuracy. In contrast, higher accuracy values under the 1-pixel, 3-pixel, and 5-pixel thresholds demonstrate stronger correspondence estimation capability.

\noindent
For the fusion task, we adopt mutual information (MI)~\cite{qu2002information}, standard deviation (SD)~\cite{ma2019infrared}, visual information fidelity (VIF)~\cite{sheikh2005information}, edge information preservation metric $Q^{AB/F}$~\cite{xydeas2000objective}, and structural similarity index measure (SSIM)~\cite{wang2004image} as evaluation metrics for fused image quality. All metrics follow the higher-is-better criterion, indicating that larger values correspond to superior fusion quality.

\noindent
\textbf{SOTA Competitors:} To validate the effectiveness of the proposed method, we conduct comprehensive comparisons with state-of-the-art registration and fusion approaches across diverse datasets. For registration, we compare our method with representative state-of-the-art approaches, including GLUNet~\cite{truong2020glu}, DiffMatch~\cite{nam2024diffusion}, and CRFT~\cite{liu2026crft}. For fusion, the comparison methods include U2Fusion~\cite{xu2020u2fusion}, TC-MoA~\cite{zhu2024task} and LUT-Fuse~\cite{yi2025lut}. Furthermore, we evaluate against recent unified registration–fusion frameworks, including UMF-CMGR~\cite{wang2022unsupervised}, MURF~\cite{xu2023murf}, C2RF~\cite{tang2025c2rf}, AUNet~\cite{lu2025net}, PGMR~\cite{zheng2025plug}, and CPMFusion~\cite{xiao2026cpmfusion}. To address the absence of enhancement capability in the compared methods, we adopt BaryIR~\cite{tang2026learning}, retrained on the corresponding datasets, as a pre-enhancement strategy.

\subsection{Comparative Experiments}

\noindent
\textbf{Qualitative Results on Image Registration:} To better visualize the registration accuracy, we interleave the registered and restored infrared and visible images. Note that the input infrared and visible images for registration contain degradations rather than being clean images. We validate the proposed method on the MFNet and FMB datasets and conduct generalization experiments on the LLVIP dataset. The qualitative experimental results are reported in Fig.~\ref{reg_qual}. It can be observed that, under degradations including low illumination, haze, stripe noise, rain, and random noise in the input images, different methods demonstrate varying adaptation capabilities. In contrast, the proposed Diff-RF method achieves relatively good performance across different scenarios. In the first scenario with low-light degradation, the source images suffer from severe spatial misalignment, and the low illumination further challenges the registration process. The proposed method achieves more accurate registration, effectively aligning targets such as pedestrians. In contrast, GLUNet, CRFT, UMF-CMGR, and C2RF exhibit obvious residual misalignment. Although DiffMatch, AUNet, PGMR, and CPMFusion mitigate part of the misalignment, they still struggle to achieve precise alignment of local details and fine structures. In the second scenario with haze degradation, we evaluate the registration results of three severely misaligned pedestrian targets and vehicles. GLUNet, DiffMatch, and PGMR can only achieve coarse alignment, with apparent deviations remaining in local contours. Meanwhile, CRFT, UMF-CMGR, C2RF, AUNet, and CPMFusion fail to effectively recover the spatial correspondence, resulting in registration failures. By contrast, Diff-RF accurately corrects the misalignment of the pedestrian targets and vehicles, achieving more complete and precise cross-modal alignment. In the third scenario, the infrared images are severely affected by stripe noise. Almost all compared methods experience performance degradation or even registration failure under such strong infrared degradation. Benefiting from the collaborative enhancement among degradation restoration, registration, and fusion, Diff-RF still maintains robust and accurate registration performance. Similarly, the fourth and fifth scenarios further consider challenging degradations, including rain and random noise, to evaluate the ability of Diff-RF to register fine-grained structures, such as lamp posts and stairs. Across all scenarios, the proposed method consistently demonstrates clear advantages over competing approaches. Moreover, in the sixth scenario for generalization evaluation, Diff-RF also exhibits strong generalization capability, accurately aligning road boundaries and achieving more precise spatial registration.

\noindent
\textbf{Quantitative Results on Image Registration:} The corresponding quantitative experimental results are reported in Tabs.~\ref{MFNet_reg_quan}-\ref{LLVIP_reg_quan}. It can be observed that our method achieves the best spatial registration performance on all validation and generalization datasets. On the MFNet dataset, the proposed method achieves the best performance in terms of EPE and estimation accuracy under the 1-pixel, 3-pixel, and 5-pixel thresholds. Specifically, compared with the second-best method PGMR, it reduces the EPE by 1.005 and improves the estimation accuracy by 0.980, 5.100, and 7.618 under the 1-pixel, 3-pixel, and 5-pixel thresholds, respectively. Similarly, on the FMB dataset, Diff-RF benefits from its degradation-aware design for complex degradation scenarios and achieves superior performance over the second-best method PGMR. Specifically, it reduces the EPE by 1.889 and improves the estimation accuracy by 1.783, 11.924, and 17.668 under the 1-pixel, 3-pixel, and 5-pixel thresholds. For the generalization evaluation on the LLVIP dataset, the proposed Diff-RF remains highly competitive and achieves the best results across all metrics. These results demonstrate its superior capability in accurately estimating flow fields and establishing precise spatial correspondences.

\begin{table}[!t]
    \renewcommand{\arraystretch}{1.2}
    \centering
    \caption{Quantitative comparison of image registration on the MFNet dataset. (\textbf{Bold}: optimal performance)}
    \begin{tabularx}{0.5\textwidth}{
    c |
    >{\centering\arraybackslash}X
    >{\centering\arraybackslash}X
    >{\centering\arraybackslash}X
    >{\centering\arraybackslash}X
    }
    \toprule
    \multicolumn{1}{c|}{\multirow{2}{*}{\textbf{Methods}}} & \multicolumn{4}{c}{\textbf{Registration on the MFNet Dataset}} \\\cline{2-5}
    & EPE & 1px & 3px & 5px \\
    \midrule
    GLUNet    & 15.686         & 2.653          & 19.432          & 38.057          \\
    DiffMatch & 19.438         & 0.809          & 6.778           & 16.419          \\
    CRFT      & 21.772         & 0.123          & 1.061           & 2.908           \\
    UMF-CMGR  & 22.810         & 0.118          & 1.042           & 2.823           \\
    MURF      & 79.089         & 0.358          & 3.064           & 7.729           \\
    C2RF      & 18.847         & 0.515          & 4.387           & 10.502          \\
    AUNet     & 10.376         & 1.694          & 13.604          & 30.364          \\
    PGMR      & 5.767          & 3.894          & 28.304          & 54.767          \\
    CPMFusion & 10.296         & 1.122          & 9.449           & 23.122          \\
    \rowcolor{pink!40}
    \textbf{Diff-RF (ours)}  &  \textbf{4.762} & \textbf{4.874} & \textbf{33.404} & \textbf{62.385} \\
    \bottomrule
    \end{tabularx}
    \label{MFNet_reg_quan}
\end{table}

\begin{table}[!t]
    \renewcommand{\arraystretch}{1.2}
    \centering
    \caption{Quantitative comparison of image registration on the FMB dataset. (\textbf{Bold}: optimal performance)}
    \begin{tabularx}{0.5\textwidth}{
    c |
    >{\centering\arraybackslash}X
    >{\centering\arraybackslash}X
    >{\centering\arraybackslash}X
    >{\centering\arraybackslash}X
    }
    \toprule
    \multicolumn{1}{c|}{\multirow{2}{*}{\textbf{Methods}}} & \multicolumn{4}{c}{\textbf{Registration on the FMB Dataset}} \\\cline{2-5}
    & EPE & 1px & 3px & 5px \\
    \midrule
    GLUNet    & 18.010         & 1.798          & 14.505          & 32.339          \\
    DiffMatch & 19.495         & 1.084          & 8.895           & 20.442          \\
    CRFT      & 27.377         & 0.069          & 0.645           & 1.840           \\
    UMF-CMGR  & 29.243         & 0.075          & 0.666           & 1.837           \\
    MURF      & 160.407        & 0.265          & 2.261           & 5.820           \\
    C2RF      & 25.488         & 0.430          & 3.467           & 7.972           \\
    AUNet     & 15.516         & 1.106          & 8.879           & 20.922          \\
    PGMR      & 7.491          & 1.763          & 14.621          & 35.038          \\
    CPMFusion & 15.038         & 0.725          & 6.117           & 15.257          \\
    \rowcolor{pink!40}
    \textbf{Diff-RF (ours)}  &  \textbf{5.602} & \textbf{3.546} & \textbf{26.545} & \textbf{52.706} \\
    \bottomrule
    \end{tabularx}
    \label{FMB_reg_quan}
\end{table}

\begin{table}[!t]
    \renewcommand{\arraystretch}{1.2}
    \centering
    \caption{Quantitative comparison of image registration on the LLVIP dataset for generalization evaluation. (\textbf{Bold}: optimal performance)}
    \begin{tabularx}{0.5\textwidth}{
    c |
    >{\centering\arraybackslash}X
    >{\centering\arraybackslash}X
    >{\centering\arraybackslash}X
    >{\centering\arraybackslash}X
    }
    \toprule
    \multicolumn{1}{c|}{\multirow{2}{*}{\textbf{Methods}}} & \multicolumn{4}{c}{\textbf{Registration on the LLVIP Dataset}} \\\cline{2-5}
    & EPE & 1px & 3px & 5px \\
    \midrule
    GLUNet    & 12.495         & 3.271          & 21.872          & 41.753          \\
    DiffMatch & 12.920         & 1.670          & 12.732          & 27.293          \\
    CRFT      & 23.731         & 0.100          & 0.823           & 2.240           \\
    UMF-CMGR  & 23.344         & 0.113          & 1.106           & 2.799           \\
    MURF      & 82.499         & 0.807          & 6.446           & 14.971          \\
    C2RF      & 20.310         & 0.450          & 3.325           & 7.518           \\
    AUNet     & 9.054          & 2.213          & 16.497          & 35.245          \\
    PGMR      & 6.371          & 2.944          & 22.942          & 48.595          \\
    CPMFusion & 10.062         & 1.117          & 9.519           & 23.223          \\
    \rowcolor{pink!40}
    \textbf{Diff-RF (ours)}    &  \textbf{5.182} & \textbf{3.649} & \textbf{28.400} & \textbf{59.140} \\
    \bottomrule
    \end{tabularx}
    \label{LLVIP_reg_quan}
\end{table}

\begin{figure*}[!t]
\centering
\includegraphics[width=1\linewidth]{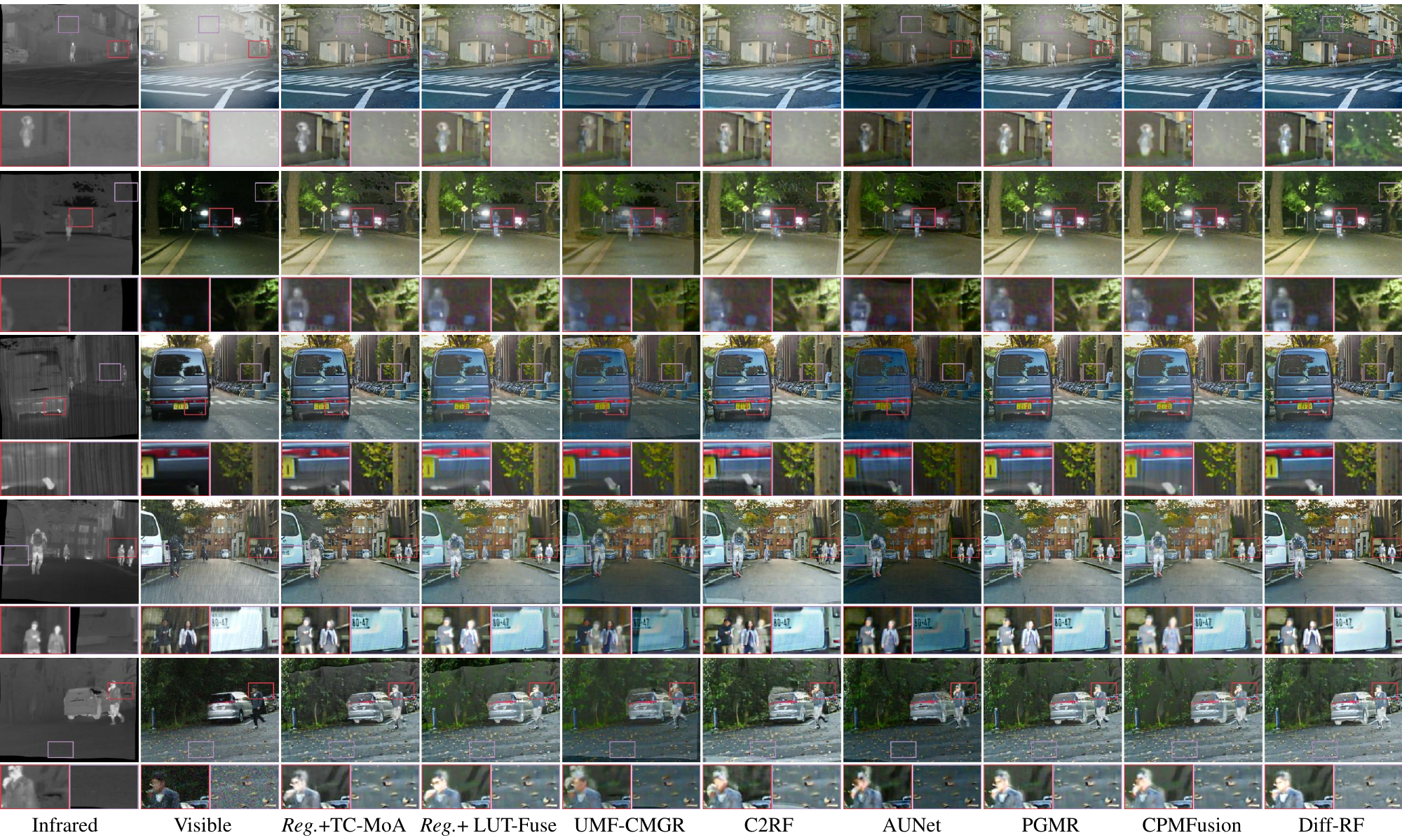}
\caption{Qualitative comparison on the MFNet dataset with state-of-the-art image registration and fusion methods. ``Reg.'' indicates that registration preprocessing is performed using the best-performing registration method on the validation set.}
\label{MFNet_fus_qual}
\end{figure*}

\begin{figure*}[!t]
\centering
\includegraphics[width=1\linewidth]{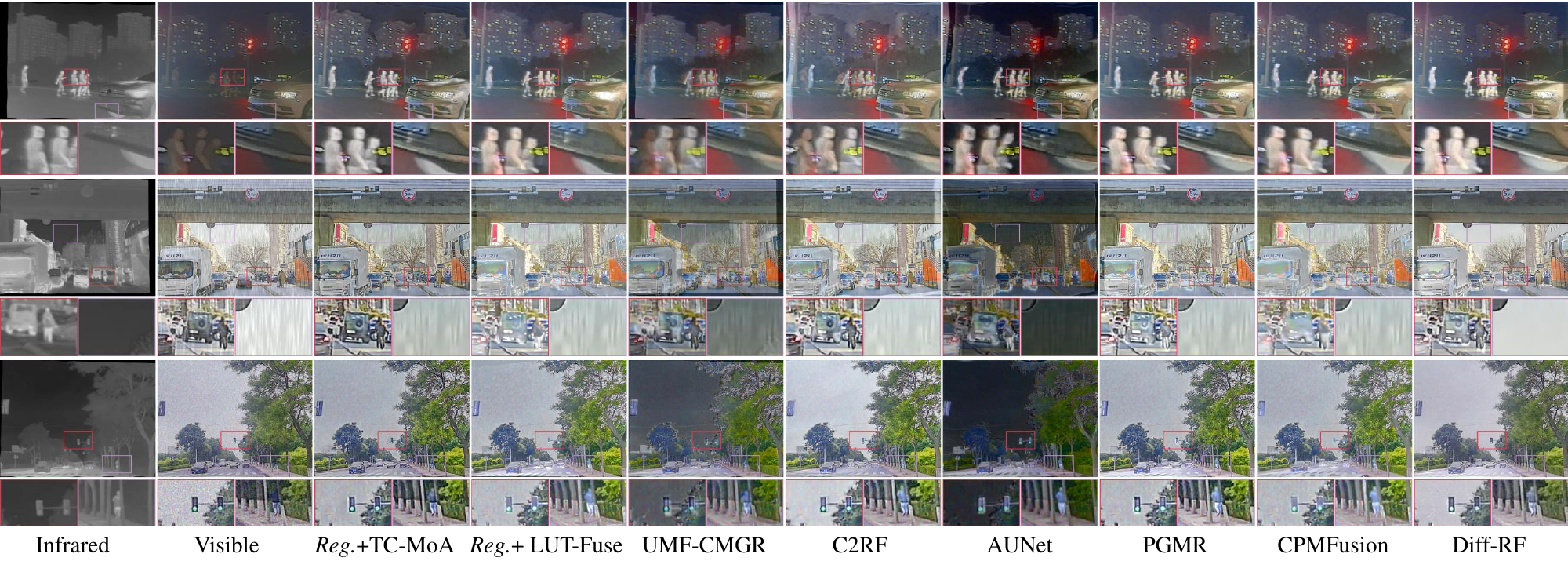}
\caption{Qualitative comparison on the FMB dataset with state-of-the-art image registration and fusion methods. ``Reg.'' indicates that registration preprocessing is performed using the best-performing registration method on the validation set.}
\label{FMB_fus_qual}
\end{figure*}

\begin{figure*}[!t]
\centering
\includegraphics[width=1\linewidth]{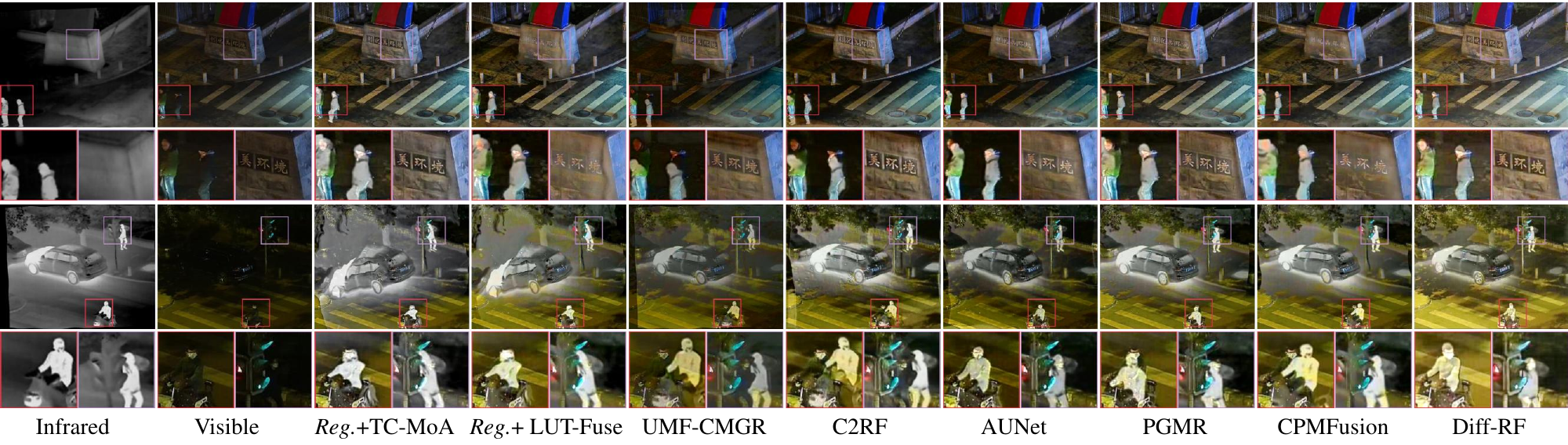}
\caption{Qualitative comparison on the LLVIP dataset for generalization evaluation with state-of-the-art image registration and fusion methods. ``Reg.'' denotes the registration preprocessing performed using the best-performing registration method selected on the validation set.}
\label{LLVIP_fus_qual}
\end{figure*}

\noindent
\textbf{Qualitative Results on Image Fusion:} For unregistered image fusion under complex degraded conditions, the qualitative comparisons are presented in Figs.~\ref{MFNet_fus_qual} and ~\ref{FMB_fus_qual}. For methods without built-in registration capability, we adopt the best-performing registration method among the comparison approaches, GLUNet, as a preprocessing step for registration before fusion. It can be observed that the proposed method exhibits three significant advantages. First, benefiting from its degradation-aware design tailored to complex scenarios, the proposed method preserves structural consistency between infrared and visible modalities under degraded conditions, effectively mitigating edge artifacts and target shifts caused by the propagation of registration errors. In the first to fourth scenarios of Fig.~\ref{MFNet_fus_qual} and first scenario of Fig.~\ref{FMB_fus_qual}, degradations including low illumination, haze, stripe noise, and rain impair the cross-modal correspondence estimation, causing UMF-CMGR, C2RF, AUNet, and CPMFusion to suffer from different degrees of registration error propagation. Consequently, their fused results exhibit target-edge artifacts, blurred details, and structural misalignment. Second, the proposed method effectively alleviates visual distortions introduced by degradations, resulting in more natural and well-balanced fused images. In first scenario of Fig.~\ref{MFNet_fus_qual}, almost all compared methods are degraded by haze and low-contrast infrared information, resulting in compromised visual quality. In contrast, the proposed method achieves clearer representations of fences and leaves, with improved detail preservation and more natural color appearance. Similarly, in the third scenario of Fig.~\ref{MFNet_fus_qual}, TC-MoA, LUT-Fuse, C2RF, AUNet, PGMR, and CPMFusion fail to effectively suppress the stripe noise from infrared images, leaving severe degradations in the fused results and affecting the representation quality. In contrast, the proposed method effectively suppresses stripe noise while preserving salient infrared contrast. In the rainy and noisy conditions, as shown in the fourth and fifth scenarios of Fig.~\ref{MFNet_fus_qual} and the second and third scenarios of Fig.~\ref{FMB_fus_qual}, TC-MoA, LUT-Fuse, UMF-CMGR, C2RF, PGMR, and CPMFusion fail to completely alleviate rain degradation in fused image, leaving noticeable rain streaks in the fused results. Although AUNet achieves relatively better suppression of rain degradation, its overall scene representation remains less natural. Moreover, almost all compared methods struggle to handle noise effectively, either introducing severe artifacts or failing to preserve clear texture details. Third, the proposed method effectively integrates salient infrared target information with rich visible details, resulting in fused images with clearer target contours, more abundant scene details, and enhanced target visibility. In the first, second, and fifth scenarios of Fig.~\ref{MFNet_fus_qual}, almost all comparison methods fail to clearly preserve the colors and decorative patterns of pedestrians' clothing and the texture details of the backpack. In contrast, the proposed method simultaneously retains salient thermal radiation information and clear visible texture and color details. Similarly, in challenging generalization scenarios shown in Fig.~\ref{LLVIP_fus_qual}, the proposed method maintains superior cross-modal fusion performance in terms of target consistency, information effectiveness, and visual naturalness. It effectively alleviates misalignment caused by complex degradations, preserves structural integrity of salient targets, suppresses degradation artifacts, and fully exploits complementary infrared and visible information, resulting in clearer and more natural fusion results.

\begin{table}[!t]
    \renewcommand{\arraystretch}{1.2}
    \centering
    \caption{Quantitative comparison of image fusion in degradation conditions on the MFNet dataset. (\textbf{Bold}: optimal performance)}
    \begin{tabularx}{0.5\textwidth}{
    c |
    >{\centering\arraybackslash}X
    >{\centering\arraybackslash}X
    >{\centering\arraybackslash}X
    >{\centering\arraybackslash}X
    >{\centering\arraybackslash}X
    }
    \toprule
    \multicolumn{1}{c|}{\multirow{2}{*}{\textbf{Methods}}} & \multicolumn{5}{c}{\textbf{Fusion on the MFNet Dataset}} \\\cline{2-6}
    & MI & SD & VIF & $Q^{AB/F}$ & SSIM \\
    \midrule
    \textit{Reg.}+U2Fusion & 1.1740          & 41.5359          & 0.4268          & 0.4244          & 0.6250          \\
    \textit{Reg.}+TC-MoA   & 1.5280          & 45.3212          & 0.5140          & 0.4804          & 0.6799          \\
    \textit{Reg.}+LUT-Fuse & 1.5326          & 44.2125          & 0.4588          & 0.4207          & 0.6257          \\
    UMF-CMGR      & 1.0782          & 31.3388          & 0.3400          & 0.2628          & 0.4581          \\
    MURF          & 0.9807          & 26.4109          & 0.2603          & 0.2033          & 0.4183          \\
    C2RF          & 1.3961          & 46.4450          & 0.3908          & 0.3963          & 0.5787          \\
    AUNet         & 1.0585          & 27.8909          & 0.3096          & 0.2086          & 0.4320          \\
    PGMR          & 1.6207          & 46.0841          & 0.5172          & 0.4835          & 0.6899          \\
    CPMFusion     & 1.5780          & 43.3107          & 0.4595          & 0.4336          & 0.6166          \\
    \rowcolor{pink!40}
    \textbf{Diff-RF (ours)}       & \textbf{1.6788} & \textbf{47.4855} & \textbf{0.5416} & \textbf{0.5041} & \textbf{0.7042} \\
    \bottomrule
    \end{tabularx}
    \label{MFNet_fus_quan}
\end{table}

\begin{table}[!t]
    \renewcommand{\arraystretch}{1.2}
    \centering
    \caption{Quantitative comparison of image fusion in degradation conditions on the FMB dataset. (\textbf{Bold}: optimal performance)}
    \begin{tabularx}{0.5\textwidth}{
    c |
    >{\centering\arraybackslash}X
    >{\centering\arraybackslash}X
    >{\centering\arraybackslash}X
    >{\centering\arraybackslash}X
    >{\centering\arraybackslash}X
    }
    \toprule
    \multicolumn{1}{c|}{\multirow{2}{*}{\textbf{Methods}}} & \multicolumn{5}{c}{\textbf{Fusion on the FMB Dataset}} \\\cline{2-6}
    & MI & SD & VIF & $Q^{AB/F}$ & SSIM \\
    \midrule
    \textit{Reg.}+U2Fusion & 1.3582          & 36.2418          & 0.4264          & 0.5051          & 0.5195          \\
    \textit{Reg.}+TC-MoA   & 1.6324          & 40.6151          & 0.5400          & 0.5613          & 0.5733          \\
    \textit{Reg.}+LUT-Fuse & 1.5820          & 39.2422          & 0.4444          & 0.4792          & 0.5334          \\
    UMF-CMGR      & 1.2039          & 32.1876          & 0.3489          & 0.3746          & 0.4363          \\
    MURF          & 1.1032          & 26.2030          & 0.2008          & 0.2588          & 0.3416          \\
    C2RF          & 1.4910          & 41.1078          & 0.4171          & 0.5027          & 0.5144          \\
    AUNet         & 1.2588          & 31.0200          & 0.2906          & 0.2498          & 0.3422          \\
    PGMR          & 1.6307          & 41.3777          & 0.5240          & 0.5482          & 0.5790          \\
    CPMFusion     & 1.6233          & 38.3419          & 0.4411          & 0.4786          & 0.5229          \\
    \rowcolor{pink!40}
    \textbf{Diff-RF (ours)}       & \textbf{2.3834} & \textbf{44.5935} & \textbf{0.5900} & \textbf{0.6049} & \textbf{0.6306} \\
    \bottomrule
    \end{tabularx}
    \label{FMB_fus_quan}
\end{table}

\begin{table}[!t]
    \renewcommand{\arraystretch}{1.2}
    \centering
    \caption{Quantitative comparison of image fusion for generalization evaluation in degradation conditions on the LLVIP dataset. (\textbf{Bold}: optimal performance)}
    \begin{tabularx}{0.5\textwidth}{
    c |
    >{\centering\arraybackslash}X
    >{\centering\arraybackslash}X
    >{\centering\arraybackslash}X
    >{\centering\arraybackslash}X
    >{\centering\arraybackslash}X
    }
    \toprule
    \multicolumn{1}{c|}{\multirow{2}{*}{\textbf{Methods}}} & \multicolumn{5}{c}{\textbf{Fusion on the LLVIP Dataset}} \\\cline{2-6}
    & MI & SD & VIF & $Q^{AB/F}$ & SSIM \\
    \midrule
    \textit{Reg.}+U2Fusion & 0.9577          & 38.5195          & 0.3249          & 0.3492          & 0.5800          \\
    \textit{Reg.}+TC-MoA   & 1.2084          & 49.8302          & 0.4375          & 0.4803          & 0.6932          \\
    \textit{Reg.}+LUT-Fuse & 1.2707          & 49.1115          & 0.3438          & 0.3642          & 0.5582          \\
    UMF-CMGR      & 0.8076          & 36.5192          & 0.2487          & 0.2701          & 0.4622          \\
    MURF          & 0.7388          & 26.4526          & 0.1592          & 0.1554          & 0.3581          \\
    C2RF          & 0.9989          & \textbf{51.6695} & 0.2873          & 0.3709          & 0.5458          \\
    AUNet         & 0.9312          & 38.8032          & 0.2697          & 0.2216          & 0.4853          \\
    PGMR          & 1.3613          & 49.5726          & 0.4336          & 0.4817          & 0.7123          \\
    CPMFusion     & 1.2898          & 48.3769          & 0.3463          & 0.3858          & 0.5550          \\
    \rowcolor{pink!40}
    \textbf{Diff-RF (ours)}       & \textbf{1.5391} & 48.5692          & \textbf{0.5151} & \textbf{0.5313} & \textbf{0.7947} \\
    \bottomrule
    \end{tabularx}
    \label{LLVIP_fus_quan}
\end{table}

\noindent
\textbf{Quantitative Results on Image Fusion:} The quantitative comparisons are presented in Tabs.~\ref{MFNet_fus_quan}-\ref{LLVIP_fus_quan}. The proposed method overall achieves superior performance across all evaluation metrics, demonstrating its capability in complementary information preservation, visual fidelity, and structural consistency. On the MFNet dataset, the proposed method achieves superior performance over the second-best method PGMR across all evaluation metrics. Specifically, it improves MI, SD, VIF, $Q^{AB/F}$, and SSIM by 0.0581, 1.4014, 0.0244, 0.0206, and 0.0143, respectively. These improvements indicate that the proposed method is more effective in preserving complementary information from infrared and visible modalities, while simultaneously enhancing structural integrity and visual fidelity of the fused images. Similarly, on the FMB dataset, the proposed method achieves the best overall performance among all compared methods. Compared with the second-ranked TC-MoA, it improves MI, SD, VIF, $Q^{AB/F}$, and SSIM by 0.7510, 3.9784, 0.0500, 0.0436, and 0.0573, respectively. Notably, the remarkable gains in VIF, $Q^{AB/F}$, and SSIM demonstrate that the proposed method effectively maintains visual information fidelity and target structural integrity under complex scenarios. Furthermore, the considerable improvement in SD reflects richer information representation and stronger contrast characteristics in the fused images. For the generalization experiment on the LLVIP dataset, as shown in Tab.~\ref{LLVIP_fus_quan}, the proposed method also achieves advanced performance. Specifically, it ranks first in four metrics, including MI, VIF, $Q^{AB/F}$, and SSIM, while achieving competitive performance in terms of SD. These results further validate the superiority of the proposed method in preserving cross-modal information and generating high-quality fused images under diverse degradation scenarios.

\subsection{Downstream Task Performance Validation}
To validate the performance of the proposed method on downstream high-level vision tasks, we conduct a series of experiments, including semantic segmentation and object detection.

\begin{figure*}[!t]
\centering
\includegraphics[width=1\linewidth]{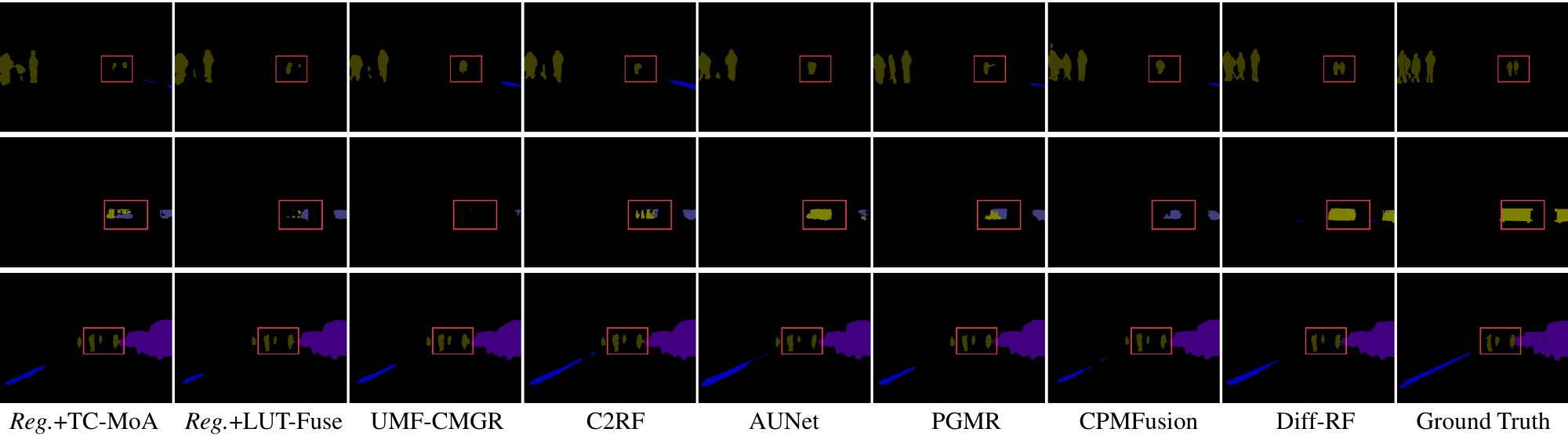}
\caption{Qualitative comparison of segmentation visualization results on the MNet dataset with state-of-the-art image fusion methods.}
\label{MFNet_seg_qual}
\end{figure*}

\noindent
\textbf{Semantic Segmentation:} For the semantic segmentation task, we employ SegFormer-B2~\cite{xie2021segformer} as the backbone network and conduct independent training on MFNet for each fused result generated by different fusion methods. The qualitative and quantitative comparisons are presented in Fig.~\ref{MFNet_seg_qual} and Tab.~\ref{MFNet_seg_quan}, respectively. For the qualitative comparison, in the first group of examples, UMF-CMGR, C2RF, AUNet, and CPMFusion exhibit evident under-segmentation, with missing human targets. Although TC-MoA, LUT-Fuse, and PGMR correctly detect the target instances, their segmentation regions are still incomplete. By contrast, Diff-RF achieves the most complete segmentation results, demonstrating superior preservation of target structures. In the second group, most compared methods exhibit under-segmentation and erroneous segmentation, whereas the proposed method achieves more accurate and complete segmentation results. Similarly, in the third group, most compared methods suffer from false human segmentation and incomplete vehicle segmentation. In contrast, the proposed method achieves segmentation results that are closest to the ground truth. For the quantitative comparison, the proposed Diff-RF achieves the best mAcc and mIoU scores, outperforming the second-best method CPMFusion in mAcc by 0.71 and surpassing the second-best method U2Fusion in mIoU by 1.49. It demonstrates the superiority of the proposed method in semantic segmentation performance.

\begin{table*}[!h]
    \caption{Quantitative results of semantic segmentation validation for all categories on the MFNet dataset. (\textbf{Bold}: optimal performance)}
    \renewcommand{\arraystretch}{1.15}
    \centering
    \resizebox{0.98\textwidth}{!}{
    \begin{tabularx}{\linewidth}{
        @{}>{\centering\arraybackslash}p{2.0cm}|
        *{9}{>{\centering\arraybackslash}X}|
        >{\centering\arraybackslash}p{1.2cm}|
        >{\centering\arraybackslash}p{1.2cm}@{}
    }
    \toprule
    \textbf{Methods} & Bac. & Car & Per. & Bike & Curve & Car\_stop & Guar. & Col. & Bump & mAcc & mIoU \\
    \midrule
    \textit{Reg.}+U2Fusion & 97.83 & 87.90 & 65.95 & 68.23 & 29.72 & 30.98 & 1.05 & 43.03 & 32.55 & 58.05          & 50.80          \\
    \textit{Reg.}+TC-MoA   & 97.90 & 88.17 & 65.80 & 68.08 & 31.03 & 26.37 & 1.75 & 43.86 & 28.13 & 57.18          & 50.12          \\
    \textit{Reg.}+LUT-Fuse & 97.81 & 87.05 & 66.50 & 66.94 & 27.70 & 23.00 & 0.84 & 42.48 & 20.99 & 54.29          & 48.14          \\
    UMF-CMGR      & 97.74 & 86.34 & 63.94 & 66.06 & 30.78 & 25.43 & 0.00 & 42.41 & 22.93 & 54.16          & 48.40          \\
    MURF          & 97.69 & 86.59 & 61.51 & 65.86 & 29.08 & 29.92 & 0.00 & 42.51 & 24.56 & 54.57          & 48.63          \\
    C2RF          & 97.83 & 88.21 & 63.28 & 66.59 & 29.58 & 34.88 & 0.38 & 43.57 & 25.51 & 56.27          & 49.98          \\
    AUNet         & 97.88 & 88.14 & 66.39 & 68.35 & 31.09 & 30.89 & 0.00 & 43.84 & 23.53 & 56.78          & 50.01          \\
    PGMR          & 97.92 & 88.20 & 68.03 & 69.06 & 30.21 & 33.65 & 0.01 & 43.10 & 19.62 & 56.11          & 49.99          \\
    CPMFusion     & 97.87 & 87.55 & 65.72 & 67.66 & 31.69 & 28.32 & 4.79 & 42.94 & 28.02 & 58.09          & 50.51          \\
    \rowcolor{pink!40}
    \textbf{Diff-RF (ours)}       &  98.01 & 88.27 & 68.70 & 68.62 & 36.25 & 31.96 & 0.52 & 42.89 & 35.38 & \textbf{58.80} & \textbf{52.29} \\
    \bottomrule
    \end{tabularx}
    }
    \label{MFNet_seg_quan}
\end{table*}

\begin{figure}[!t]
\centering
\includegraphics[width=1\linewidth]{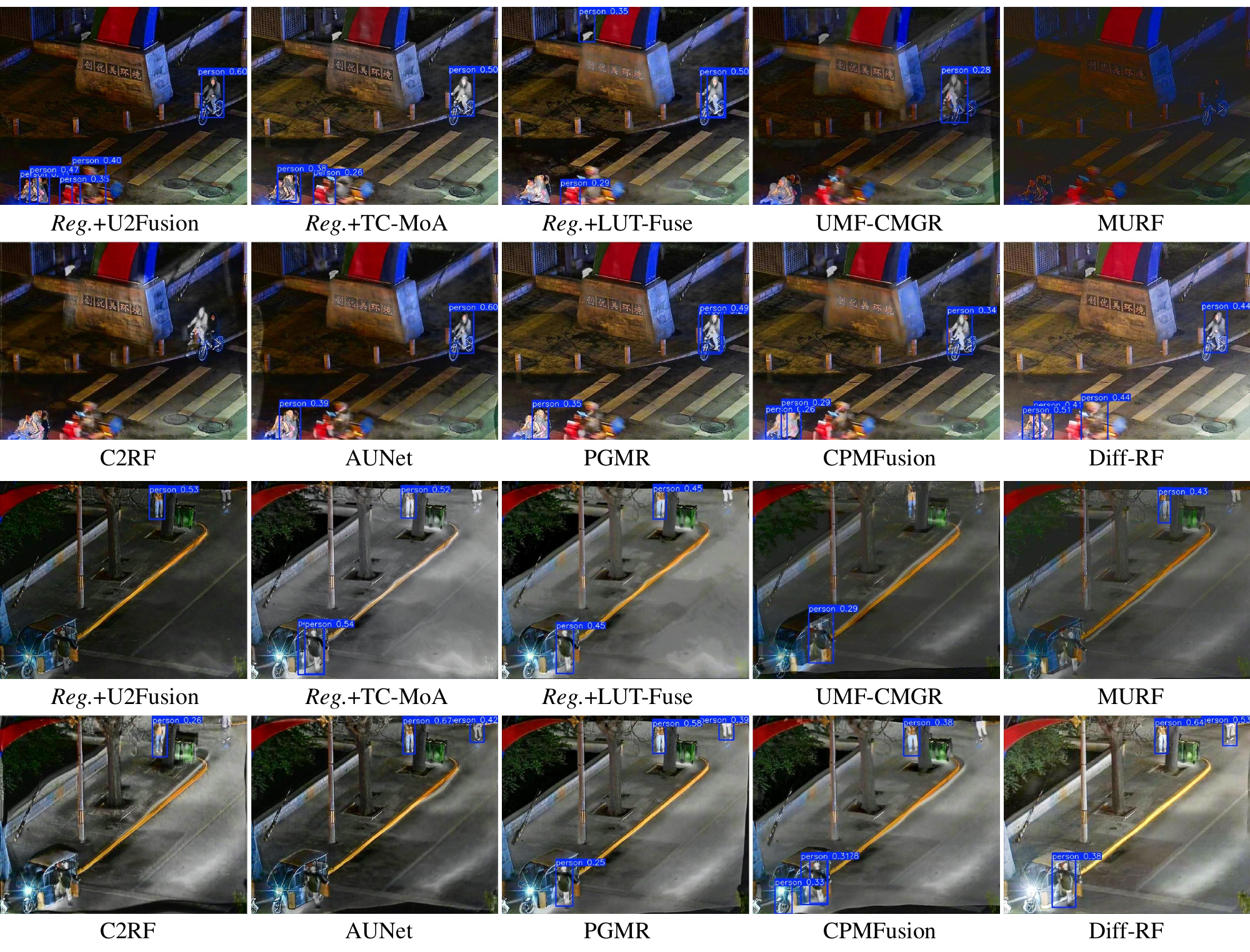}
\caption{Qualitative comparison of object detection results on the LLVIP dataset with state-of-the-art image fusion methods.}
\label{LLVIP_det_qual}
\end{figure}

\noindent
\textbf{Object Detection:} For the object detection task, we adopt YOLOv11\footnote{\url{https://github.com/ultralytics/ultralytics}} as the detector on the LLVIP dataset and initialize it with pretrained weights. For each fusion method, the detector is retrained independently on the corresponding dataset to evaluate the detection performance. The qualitative and quantitative comparisons are presented in Fig.~\ref{LLVIP_det_qual} and Tab.~\ref{LLVIP_det_quan}, respectively. For the qualitative comparison, in the first group of examples, TC-MoA, LUT-Fuse, UMF-CMGR, MURF, C2RF, AUNet, PGMR, and CPMFusion suffer from evident missed detections. Although U2Fusion successfully detects all targets, it introduces false positives. By contrast, the proposed Diff-RF achieves more complete and accurate detection results. Similarly, in the second group of examples, all compared methods except PGMR exhibit missed detections. Although PGMR successfully detects the targets, its detection confidence remains relatively low. Overall, Diff-RF achieves more accurate detection results with higher confidence scores. For the quantitative evaluation, the proposed method achieves the highest Recall and mAP@50 values, demonstrating the best overall detection performance among all compared methods.

\begin{table}[!t]
    \renewcommand{\arraystretch}{1.2}
    \centering
    \caption{Quantitative comparison of object detection on the LLVIP dataset. (\textbf{Bold}: optimal performance)}
    \begin{tabularx}{0.48\textwidth}{
    c
    >{\centering\arraybackslash}X
    >{\centering\arraybackslash}X
    >{\centering\arraybackslash}X
    >{\centering\arraybackslash}X
    }
    \toprule
    \textbf{Methods} & Precision & Recall & mAP@50 & mAP@50:95 \\
    \midrule
    \textit{Reg.}+U2Fusion & 0.802          & 0.605          & 0.706          & 0.323          \\
    \textit{Reg.}+TC-MoA   & 0.764          & 0.654          & 0.745          & 0.322          \\
    \textit{Reg.}+LUT-Fuse & 0.755          & 0.603          & 0.709          & 0.312          \\
    UMF-CMGR      & 0.507          & 0.441          & 0.441          & 0.147          \\
    MURF          & 0.595          & 0.447          & 0.469          & 0.178          \\
    C2RF          & 0.624          & 0.483          & 0.553          & 0.194          \\
    \rowcolor{pink!40}
    AUNet         & \textbf{0.845}          & 0.677          & 0.796          & \textbf{0.362}          \\
    PGMR          & 0.833          & 0.686          & 0.785          & 0.340          \\
    CPMFusion     & 0.769          & 0.605          & 0.702          & 0.273          \\
    \rowcolor{pink!40}
    \textbf{Diff-RF (ours)}       &  0.821 & \textbf{0.731} & \textbf{0.800} & 0.359 \\
    \bottomrule
    \end{tabularx}
    \label{LLVIP_det_quan}
\end{table}

\subsection{Ablation Experiments}
To validate the improvements brought by the proposed ideas and investigate the contribution of each module to the overall performance, we conduct a series of ablation experiments on the MFNet dataset. We evaluate the variants by the model removing IRM (w/o IRM), disabling the overall collaborative training strategy (w/o CTS), and the model removing fusion feature-guided registration (w/o FGR), respectively. The experimental results are reported in Tabs.~\ref{ablation_quan_reg} and~\ref{ablation_quan_fus}.

For the image registration task, we evaluate registration accuracy using EPE and the estimation accuracy under different error thresholds. Removing IRM, which makes the model directly learn degradation-aware registration from degraded observations in a data-driven manner, results in notable performance degradation even after retraining. Specifically, the EPE increases by 0.622, while the 1-pixel, 3-pixel, and 5-pixel accuracy decreases by 0.714, 3.956, and 5.544, respectively. After removing fusion feature-guided registration, the model loses the visual guidance derived from fusion features and relies solely on geometric estimation. Consequently, all registration metrics, including EPE and the 1-pixel, 3-pixel, and 5-pixel accuracy, exhibit noticeable degradation. Similarly, without the collaborative training strategy, the modules in the framework operate independently, leading to performance degradation across all evaluation metrics. These results highlight the importance of jointly optimizing the entire framework rather than individually optimizing each module. These results validate the effectiveness of the underlying insight of the proposed method.

For the image fusion task, at the time, IRM is removed, all fusion metrics, including MI, SD, VIF, $Q^{AB/F}$, and SSIM, exhibit clear degradation. This demonstrates that intra-modal restoration effectively alleviates modality-specific degradations and provides more reliable representations for subsequent registration and fusion. When FGR is removed, MI, SD, VIF, and SSIM consistently decrease, demonstrating the importance of fusion-guided registration. Although $Q^{AB/F}$ shows negligible variation (0.50408 of w/o FGR vs. 0.50414 of full model), the overall performance degradation across multiple metrics verifies the effectiveness of FGR. Similarly, removing the collaborative training strategy leads to performance degradation across fusion metrics. These results highlight the importance of joint optimization and collaboration between modules for achieving effective fusion. Overall, these results confirm the necessity of each component and demonstrate that their collaborative interaction is essential for achieving robust multi-modal image fusion under complex degradation conditions.

\begin{table}[t]
    \centering
    \renewcommand{\arraystretch}{1.25}
    \caption{Quantitative comparison of the ablation experiment of the general components on MFNet. (\textbf{Bold}: optimal performance)}
    \resizebox{0.5\textwidth}{!}{
    \begin{tabular}{c@{\,~~~~}c@{\,~~~~}c@{\,~~~~}c@{\,~~~~}|c@{\,~~~~}c@{\,~~~~}c@{\,~~~~}c@{\,~~~~}}
        \toprule
        \textit{IRM} & \textit{CDRM} & \textit{CTS} & \textit{FGR} & EPE & 1px & 3px & 5px  \\
        \midrule
        \checkmark & \checkmark & \checkmark &  & 4.822 & 4.777 & 33.078 & 61.878 \\
           & \checkmark & \checkmark & \checkmark & 5.384 & 4.160 & 29.448 & 56.841 \\
         \checkmark & \checkmark &  & \checkmark & 4.799 & 4.840 & 33.066 & 61.892 \\
         \rowcolor{pink!40}
          \checkmark & \checkmark & \checkmark & \checkmark & \textbf{4.762} & \textbf{4.874} & \textbf{33.404} & \textbf{62.385} \\
        \bottomrule
    \end{tabular}
    }\label{ablation_quan_reg}
\end{table}

\begin{table}[t]
    \centering
    \renewcommand{\arraystretch}{1.25}
    \caption{Quantitative comparison of the ablation experiment of the general components on MFNet. (\textbf{Bold}: optimal performance determined by the original full-precision values when rounded values are identical)}
    \resizebox{0.5\textwidth}{!}{
    \begin{tabular}{c@{\,~~~~}c@{\,~~~~}c@{\,~~~~}c@{\,~~~~}|c@{\,~~~~}c@{\,~~~~}c@{\,~~~~}c@{\,~~~~}c@{\,~~~~}}
        \toprule
        \textit{IRM} & \textit{CDRM} & \textit{CTS} & \textit{FGR} &MI & SD & VIF & $Q^{AB/F}$ & SSIM  \\
        \midrule
        \checkmark & \checkmark & \checkmark &  & 1.6751 & 47.4262 & 0.5406 & 0.5041 & 0.7035 \\
           & \checkmark & \checkmark & \checkmark & 1.6060 & 44.7111 & 0.5182 & 0.5028 & 0.6979 \\
         \checkmark & \checkmark &  & \checkmark & 1.6587 & 47.4375 & 0.5382 & 0.5008 & 0.6996 \\
         \rowcolor{pink!40}
          \checkmark & \checkmark & \checkmark & \checkmark & \textbf{1.6788} & \textbf{47.4855} & \textbf{0.5416} & \textbf{0.5041} & \textbf{0.7042} \\
        \bottomrule
    \end{tabular}
    }\label{ablation_quan_fus}
\end{table}

\subsection{Computational Overhead}
To evaluate the computational overhead of the proposed method, we provide a comprehensive analysis in terms of parameter count, computational complexity, inference time, and GPU memory consumption. All experiments are conducted on an NVIDIA RTX 3090 GPU. The quantitative comparisons are presented in Table ~\ref{computational overhead}. Although the proposed method incorporates diffusion models, it maintains a moderate computational cost, with 129.923M parameters, 149.052G FLOPs, an inference time of 0.654s, and a GPU memory consumption of 2.593G. Overall, the computational overhead for the diffusion-based image registration and fusion method remains acceptable.

\begin{table}[t]
    \renewcommand{\arraystretch}{1.4}
    \setlength{\aboverulesep}{0.3ex}
    \setlength{\belowrulesep}{0.3ex}
    \centering
    \caption{Quantitative computational overhead comparison of Diff-RF. (Params.: number of parameters, FLOPs: floating-point operations, Time: inference time)}
    \begin{tabularx}{0.5\textwidth}{
    c
    >{\centering\arraybackslash}X
    >{\centering\arraybackslash}X
    >{\centering\arraybackslash}X
    >{\centering\arraybackslash}X
    }
    \toprule
    \textbf{Methods} & Params (M) & FLOPs (G) & Time (s) & Memory (G) \\
    \midrule
    Diff-RF & 129.923 & 149.052 & 0.654 & 2.593 \\
    \bottomrule
    \end{tabularx}
    \label{computational overhead}
\end{table}

\section{Conclusion}
In this paper, we proposed Diff-RF, a mutually reinforced image registration and fusion diffusion framework via degradation-aware learning for high-quality fusion of unregistered multi-modal images under complex degradation conditions. Given that registration and fusion under complex degraded scenarios remain insufficiently explored, Diff-RF investigates the intrinsic coupling among registration, fusion, and information restoration. First, we propose an intra-modal restoration module that exploits modality-specific information to alleviate degradations and recover registration-friendly structural representations from degraded source images. Moreover, we develop a cross-modal diffusion registration and fusion module that establishes bidirectional interaction between registration and fusion. By incorporating fusion-derived visual guidance and correspondence-based geometric constraints into the diffusion process, the proposed framework progressively estimates cross-modal correspondences and refines spatial alignment. Through the collaborative optimization of restoration, registration, and fusion, Diff-RF achieves robust performance under challenging imaging conditions. Extensive experiments on multiple extended datasets demonstrate that Diff-RF consistently outperforms state-of-the-art methods in terms of both registration accuracy and fusion quality, especially under severe degradation scenarios involving low illumination, haze, random noise, rain, and stripe noise.

\bibliographystyle{IEEEtran}
\bibliography{egbib}

\end{document}